# Shapley-based Explainable AI for Clustering Applications in Fault Diagnosis and Prognosis

Joseph Cohen[1*], Xun Huan[2], and Jun Ni[2]

[1]University of Michigan, Michigan Institute for Data Science, Ann Arbor MI 48105, USA

[2]University of Michigan, Department of Mechanical Engineering, Ann Arbor MI 48105, USA

*`cohenyo@umich.edu`, ORCID iD: 0000-0001-8932-4367

**Abstract.** Data-driven artificial intelligence models require explainability in intelligent manufacturing to streamline adoption and trust in modern industry. However, recently developed explainable artificial intelligence (XAI) techniques that estimate feature contributions on a model-agnostic level such as SHapley Additive exPlanations (SHAP) have not yet been evaluated for semi-supervised fault diagnosis and prognosis problems characterized by class imbalance and weakly labeled datasets. This paper explores the potential of utilizing Shapley values for a new clustering framework compatible with semi-supervised learning problems, loosening the strict supervision requirement of current XAI techniques. This broad methodology is validated on two case studies: a heatmap image dataset obtained from a semiconductor manufacturing process featuring class imbalance, and a benchmark dataset utilized in the 2021 Prognostics and Health Management (PHM) Data Challenge. Semi-supervised clustering based on Shapley values significantly improves upon clustering quality compared to the fully unsupervised case, deriving information-dense and meaningful clusters that relate to underlying fault diagnosis model predictions. These clusters can also be characterized by high-precision decision rules in terms of original feature values, as demonstrated in the second case study. The rules, limited to 1-2 terms utilizing original feature scales, describe 12 out of the 16 derived equipment failure clusters with precision exceeding 0.85, showcasing the promising utility of the explainable clustering framework for intelligent manufacturing applications.

## 1 Introduction

A significant limitation of existing AI techniques that threatens trust, adoption, and maturity in manufacturing industry is that it is exceedingly difficult to understand how developed models make predictions, particularly with arbitrarily deep and nonlinear neural networks. This lies at the forefront of what McKinsey & Company cited as the most significant challenge facing companies in implementing Industry 4.0 solutions in 2020 outside of the COVID-19 pandemic: a limited understanding of the technology itself (Agrawal et al. 2021). In light of these central challenges, the field of explainable artificial intelligence (XAI) has recently emerged as a research area exploring various approaches. Explainable and trustworthy data-driven methodology can improve decision-making for fault diagnosis as well as the broader field of prognostics and health management (PHM), enabling cost-saving intelligent predictive maintenance and effective resource allocation strategies (Hrnjica and Softic 2020).

XAI methods vary by aspects such as explanation scope, model specificity, and the location of the explanations (Ahmed et al. 2022). In recent years, XAI methods have been developed for several applications including manufacturing cost estimation (Yoo and Kang 2021), predictive maintenance (Hrnjica and Softic 2020; Serradilla et al. 2021), privacy-preserving industrial applications (Ogrezeanu et al. 2022), semiconductor defect classification (Lee et al. 2022), and quality management in semiconductor manufacturing (Senoner et al. 2021). Most of these applications attempt to quantify feature attributions for predictions and global feature importance, which historically have been model-specific and derived from nonparametric, decision tree-based modeling (Ahmed et al. 2022). Features are also inherently explainable for linear regression models, with the coefficients directly giving the attribution (Sofianidis et al. 2021). However, for nonlinear parametric models, learned weight values may give misleading results both due to differences in variable scale as well as the nonlinearities in the architecture. As a result, XAI methods have been recently developed to quantify feature attributions on a model-agnostic level, compatible with high-fidelity nonlinear supervised learning models of arbitrary complexity.

One prominent model-agnostic XAI technique is the Local Interpretable Model-Agnostic Explanations (LIME) method. The purpose of LIME is to explain individual black-box model predictions, accomplished by weighing perturbed data samples around a neighborhood of an observation of interest and obtaining a low-fidelity explainer model (Molnar 2022). The output explainer model is typically a weighted linear and/or sparse method that is inherently explainable as a surrogate model, recovering the local decision boundaries around the observation of interest. The advantage of LIME is its model-agnostic capability of mixing low- and high-fidelity predictive models, taking advantage of the discrimination capabilities of high-fidelity nonlinear modeling while simultaneously allowing local explanations provided by the low-fidelity explainer. This multi-fidelity approach also enables the usage of different feature sets; for example, highly accurate convolutional neural networks (CNNs) with learned feature representations can be utilized in conjunction with LASSO-regularized linear explainers trained on human-explainable features (Molnar 2022). However, there are significant limitations with LIME, including instability stemming from the sampling technique used. The resulting explanations may be brittle and only valid within the target observation's local neighborhood, which is in itself non-trivial to define (Molnar 2022).

Shapley-based techniques have been developed as an alternative to obtain both local and global feature attributions (Senoner et al. 2021). Originating from game theory economics, the Shapley value was first introduced in 1951 by Lloyd Shapley, who later won the Nobel Memorial Prize in Economic Sciences in 2012. The concept for Shapley values addresses the fundamental question: in a cooperative game, how does one determine the fairest payoff for all players, considering their individual contributions to the game? In the last decade, Shapley values have been reexplored for machine learning contexts, in which a Shapley value is interpreted as the average marginal contribution of a feature across all possible feature sets.

Computing the exact Shapley value is computationally expensive and complexity scales exponentially on the number of features, making it exceedingly difficult when data are high-dimensional. As a result, several approximation methods have been proposed to obtain Shapley-inspired feature attributions. Štrumbelj and Kononenko proposed a stochastic Monte Carlo-based sampling approach that approximates Shapley values via permuting and splicing data instances (Štrumbelj and Kononenko 2014). Another approximation technique was later developed by Lundberg *et al*. with SHapley Additive exPlanations (SHAP), a class of deterministic methods originally developed for decision tree models (TreeSHAP) that has since been extended to provide additive explanations regardless of the model (Lundberg et al. 2017).

Since the inception of the SHAP method, researchers have also explored the utility of the values beyond simply offering post-hoc model interpretations. For example, Senoner *et al*. used information from SHAP analysis in the semiconductor domain to identify key process and quality drivers, recommending candidate improvement actions to significantly reduce yield loss at Hitachi Energy in Zurich, Switzerland (Senoner et al. 2021). In addition, Cooper *et al.* detailed how SHAP values provide valuable context for supervised clustering analysis, discovering key subgroups in a COVID-19 symptomatology case study (Cooper et al. 2021). Although the clusters from Cooper *et al*.'s work are derived from a pipeline containing several nonlinear transformations such as those within the predictive model and stochastic dimensionality reduction, the resulting clusters are human-explainable and can be characterized with high precision by simple decision rules.

However, methods thus far have not proposed Shapley-based clustering analysis to include semi-supervised contexts in which only partial labeling is available to construct data-driven models, a key limitation for practical application. In addition, existing methods have not evaluated the utility of Shapley methods for model predictions with class imbalance. **This paper aims to address these research gaps by proposing a new clustering framework extensible for semi-supervised problem scenarios.** The main contributions of this paper are summarized as follows:

1. Developing a new Shapley-based clustering framework based on the level of supervision, with three cases explored: unsupervised, semi-supervised, and fully supervised clustering;
2. Utilizing Shapley-based explanations to derive useful, high-precision decision rules in the context of fault diagnosis and prognosis;
3. Validating on two diverse industrial use cases: a semiconductor manufacturing heatmap dataset featuring heavy class imbalance, and a benchmark dataset used in the PHM 2021 Data Challenge concerning turbofan engine degradation.

The first case study explores unsupervised and semi-supervised cases, whereas the second focuses on Shapley-explainable clustering for a fully supervised learning scenario. The case studies vary in data type (RGB image heatmaps versus time series data), level of supervision, as well as application area (semiconductor manufacturing versus aerospace prognostics), demonstrating the flexibility of the proposed methodology.

## 2 Methodology

This paper explores a novel explainable clustering framework. The overall objective is to obtain information-dense clusters that relate to the underlying predictions of a semi-supervised or fully supervised trained model, which can be further described with simple rules with high precision. The methodology utilizes the following tools:

1. SHAP and Monte Carlo-based stochastic sampling methods for Shapley value analysis;
2. Uniform Manifold Approximation and Projection (UMAP) for dimensionality reduction (recommended for visualization and addressing high dimensionality);
3. Hierarchical Density-Based Spatial Clustering of Applications with Noise (HDBSCAN) to derive dense clusters;
4. SkopeRules for learning linguistic, simplified, and heterogeneous cluster descriptions (recommended for tabular or inherently explainable features as opposed to stochastically derived latent feature representations).

The methodology addresses the need for explainable clustering extensible for semi-supervised fault diagnosis and prognosis problem scenarios common in manufacturing datasets. All steps will be further elaborated in this section. A block diagram of the methodology, illustrating the key components as applied for unsupervised, semi-supervised, and fully supervised case studies, is provided in **Fig. 1.**

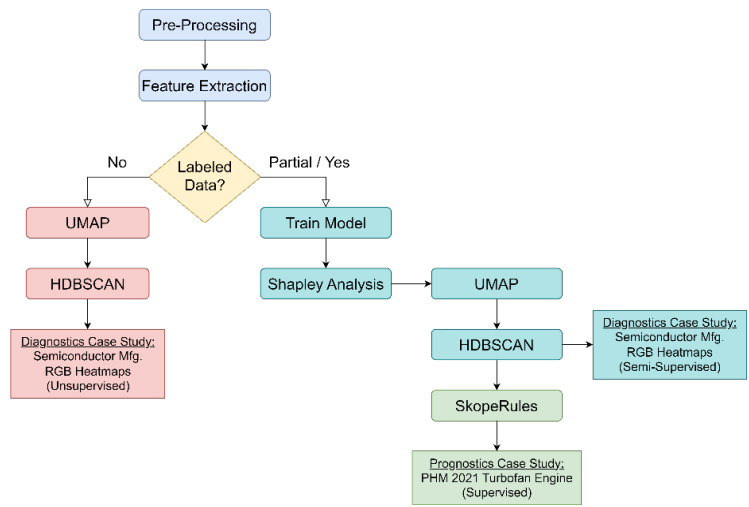

**Fig. 1** Proposed clustering methodology primarily based on UMAP and HDBSCAN techniques for unsupervised, semi-supervised, and fully supervised cases. The framework is augmented with Shapley value analysis tied to classification and/or regression model predictions when appropriate for semi-supervised and fully supervised cases

### 2.1 Shapley Value Analysis

Mathematically, the Shapley value for a feature $j$ and sample $x$ is defined in **Eq. 1** (Lundberg et al. 2017):

$$\phi_j(f, x) = \sum_{S \subseteq F \setminus \{j\}} \frac{|S|!\,(N - |S| - 1)!}{N!} [f_x(S \cup \{j\}) - f_x(S)] \quad (1)$$

in which $F$ is the power set of all features, $N$ is the total number of features, and $S$ is a subset that excludes feature $j$. $f_x(S \cup \{j\}) - f_x(S)$ is the estimated marginal contribution of adding feature $j$ to the feature subset $S$ for sample $x$, requiring the repeated evaluation of model $f$. The interpretation of the Shapley value is that it is the amount feature $j$ contributes to the prediction of sample $x$ beyond a baseline average prediction or expectation.

Local accuracy, the most essential property of Shapley theory, is defined in **Eq. 2** and states that the sum of all attributions equals the prediction:

$$f(x) = \phi_0(f) + \sum_{j=1}^{N} \phi_j(f, x) \quad (2)$$

where $\phi_0(f)$ represents the baseline prediction of the model $f$ and the local prediction $f(x)$ is explainable via a linear sum of the obtained Shapley values.

In this methodology, Shapley value analysis is used as an informative transformation of the feature space leading up to the subsequent clustering steps. The result of this is being able to explain any target prediction from a trained nonlinear predictive model as a linear combination of the baseline prediction and the localized feature attributions provided by the Shapley values, as in **Eq. 2**. However, there is further utility in Shapley values, as they can provide necessary structure to derive clusters that relate to a target prediction and contain meaningful information content (Cooper et al. 2021). Two techniques for obtaining approximations of the Shapley values are utilized in this paper: Štrumbelj and Kononenko's Monte Carlo-based sampling approach (Štrumbelj and Kononenko 2014) as well as Lundberg *et al.*'s SHAP method (Lundberg et al. 2017). In their paper, Lundberg *et al.* proved that the obtained SHAP values satisfy three important properties, justifying their usage as Shapley approximations: local accuracy (**Eq. 2**), missingness, and consistency.

### 2.2 UMAP: Uniform Manifold Approximation and Projection for Dimensionality Reduction

UMAP is a recent dimensionality reduction and visualization technique developed by McInnes *et al* (McInnes et al. 2018). Similar to t-Distributed Stochastic Neighbor Embedding (t-SNE), UMAP aims to project relationships present in a high-dimensional space onto low-dimensional embeddings using stochastically derived component spaces. To construct this projection, McInnes et al. details the construction of a fuzzy topological graph that captures essential correlations in local neighborhoods and the subsequent learning of the UMAP-embedded space (McInnes et al. 2018). McInnes *et al.* found that compared to t-SNE, UMAP better captures global behavior, and can be made flexible for several use cases depending on user-specified parameters.

UMAP revolves around three key tunable parameters: the number of neighbors to estimate the size of the local neighborhood, the minimum distance controlling the tightness of packed points, and the number of extracted UMAP components (the dimensionality of the embedding). It is important to note that unlike other standard dimensionality reduction techniques such as principal components analysis (PCA), UMAP components do not preserve original densities; therefore, the relative distances between obtained clusters may not be meaningful. However, when the number of neighbors is set appropriately high to capture global behavior (depending on application and dataset sample size) and the minimum distance is low (e.g., equal to 0), UMAP-based clustering has been empirically successful in improved visualizations that pair well with Shapley analysis for added explainability (Cooper et al. 2021).

### 2.3 HDBSCAN: Hierarchical Density-based Spatial Clustering of Applications with Noise

One of the most significant challenges with traditional clustering algorithms such as k-means, fuzzy c-means, and Gaussian mixture modeling is the need to specify the number of clusters *a priori*. Heuristics such as the "elbow method" have been employed to estimate the number of clusters as a computationally efficient alternative to other intrinsic clustering quality metrics such as silhouette score coefficients (Nainggolan et al. 2019; Shahapure and Nicholas 2020). However, this remains quite subjective, and density-based clustering approaches have since been developed to autonomously estimate the number of dense regions based on target criteria informed by specified parameters.

DBSCAN was developed as a deterministic algorithm to derive clusters based on dense data regions and depends on the epsilon parameter to represent the maximum local neighborhood distance for any two potentially neighboring data samples (Schubert et al. 2017). However, it is difficult to tune this parameter in practice. To address these shortcomings, HDBSCAN was developed as a hierarchical density-based clustering alternative that constructs a minimal spanning tree to establish the cluster hierarchy (McInnes et al. 2017). Like DBSCAN before it, HDBSCAN only clusters dense regions and excludes noisy samples. HDBSCAN is much easier for operators to tune, with two main key parameters: the minimum cluster size, and the minimum samples to be considered for noise points (in which higher settings result in more conservative clusters). Most notably, the combination of UMAP and HDBSCAN has shown success for fully unsupervised clustering of the MNIST handwritten digit benchmark, which has been a challenging research problem to date (McInnes 2018). However, the combination of UMAP and HDBSCAN has not yet been sufficiently explored for industrial big data or intelligent manufacturing applications.

**SkopeRules: Human-Explainable Cluster Descriptions**

After obtaining HDBSCAN clusterings, linguistic cluster descriptions can be obtained by fitting high-precision rules that can help "scope" a particular cluster. These rules will be learned via the SkopeRules method (Gardin et al. 2018), which learns highly discriminative rules in accordance to specified precision and recall thresholds that are then deduplicated to ensure heterogeneity. This approach is inspired by Cooper *et al.*'s subgroup discovery approach with regards to COVID-19 symptomatology, which similarly obtained rule-based descriptions to describe shared symptoms of COVID-19 positive patients (Cooper et al. 2021). However, in this paper, the approach is further constrained by providing the following three heuristics to prevent overfitting and maximize the utility and explainability provided by the rules:

1. Learned rules may only consist of the top 10 highest ranked features in mean absolute Shapley values;
2. Each rule may only have up to two terms;
3. Each rule may only be described in terms of the original feature values.

As a result, as shown in **Fig. 1**, these descriptions will only be provided for the fully supervised PHM case study to explain the cluster subgroups in terms of the physical meaning of the original tabular features.

## 3    Case Study: Semiconductor Manufacturing Heatmap Dataset

As mentioned previously, the first case study explored in this paper examines a dataset of in-process measurement profiles aggregated into RGB heatmap images. This real-world industrial application from the semiconductor manufacturing domain features 59,077 total samples, of which approximately 1000 have been identified and labeled by domain experts as faulty. Therefore, the vast majority of this dataset is unlabeled, but is mostly collected from periods of nominal operation. Therefore, the unlabeled data are assumed to be a part of one diverse "Normal" class. The faults, designated Fault 1 and Fault 2, are dataset anomalies that represent local anomalies (Fault 1), or global anomalies (Fault 2). This paper further advances the authors' previous contributions; previously, a multiclass semi-supervised anomaly detection algorithm (Cohen and Ni 2022) as well as a fuzzy clustering approach were developed and validated using this dataset (Cohen and Ni 2021). In this paper, however, we consider the utilization of Shapley-based XAI techniques to enhance clustering performance and offer classification insights for this realistic, weakly labeled, and challenging problem scenario in advanced manufacturing.

This case study is first reexplored on a purely unsupervised basis (i.e., without knowledge of available labels) to evaluate the utility of UMAP and HDBSCAN. This unsupervised clustering result is then compared to the semi-supervised case, in which the partial labels are directly used to construct supervised learning models that enable Shapley-based clusterings powered by the SHAP approximation method. Furthermore, we will provide comparisons to assess the utility of Shapley-based clustering assignments for lower-performing models. To summarize, this case study will focus on answering the following research questions: 1) How does the clustering result change when SHAP can be used versus a purely unsupervised setting? 2) How do SHAP-based clustering assignments differ when the underlying predictive model can no longer accurately predict rare anomaly types? Addressing these questions will provide fundamental insights for the utility of Shapley-based clustering extended for the class of weakly labeled semi-supervised learning problems prevalent in manufacturing industry.

### 3.1 Unsupervised Clustering with UMAP and HDBSCAN

To prepare the dataset for clustering, convolutional autoencoder (CAE)-based feature extraction and pre-processing steps detailed in past work (Cohen and Ni 2022) are first executed to reduce the dimensionality of the dataset to 20 latent and standardized features. UMAP is then used to visualize the feature space in 2 dimensions, with the number of neighbors set to 200 and minimum distance between points set to 0 to better capture global behavior amenable for clustering. Finally, HDBSCAN is executed with a minimum cluster size of 20 and a noise setting of 1800 samples to reduce the overall number of clusters. The results of this clustering are visualized in **Fig. 2**.

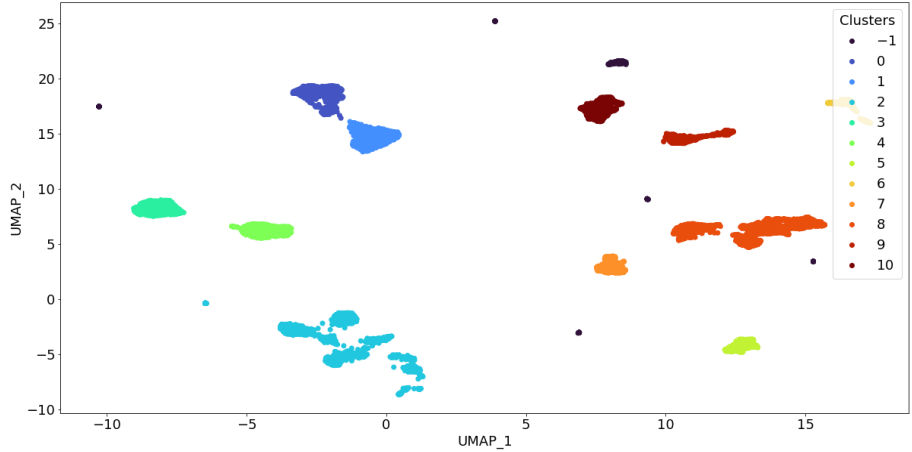

**Fig. 2** UMAP + HDBSCAN unsupervised clustering for semiconductor manufacturing heatmap case study, in which approximately 2% of the samples are identified as unclusterable noise (designated as the -1 label)

Because of the stochastic components obtained by UMAP, the clustering space and assignments differ from run-to-run, with the percentage unclustered by HDBSCAN varying from 2%-7% under these parameter settings. However, the normalized mutual information (NMI) metric, used to quantify the mutual information between assignments and reflects run-to-run stability, exceeds 0.90 between runs. The combination of UMAP and HDBSCAN for unsupervised clustering does not allow for meaningful borderline case detection due to the lack of preservation of the relative distances. The visualization capability is important for explainability and obtaining a "big picture" summary of the dataset, but the clusters themselves lack context. Importantly, this approach is not able to separate global or local anomalies into distinct clusters on a purely unsupervised basis. Therefore, this is not recommended as a technique to discover new anomaly types. Some of the unclustered samples, however, are indeed anomalous; it is recommended that in practical use, operators examine the unclustered samples for potential anomalies or unconventional data relative to the overall distribution.

### 3.2 Semi-Supervised SHAP-based Clustering

For the semi-supervised portion of this case study, the partial labels provided by domain experts are used to build two black-box data-driven models that are trained on the incomplete labeling set. This is meant to simulate the scenario in which imperfect data-driven models are deployed to classify rare events. Both models are simplified artificial neural network (ANN) models implemented using scikit-learn with a single hidden layer of 100 neurons that take the CAE-derived 20-dimensional latent feature vectors as inputs. Due to the significant class imbalance and labeling uncertainty, it is likely that these models overfit the labeled anomalies. Model 1 is characterized by CAE-extracted features normalized via z-score standardization and can classify the Fault 1 and Fault 2 anomalies with excellent precision and recall, assuming no significant labeling error. Model 2, with CAE-extracted features normalized via min-max normalization instead, is unable to classify Fault 1 to the same extent. The performance of both models are compared in **Table 1**.

**Table 1** Classification performance for Model 1 and Model 2, with both models trained on partial labels. Model 1 offers superior performance for classification

| Model | Classification | Precision | Recall | F1-Score |
|---|---|---|---|---|
| Model 1 | Normal | 1.00 | 1.00 | 1.00 |
|  | Fault 1 | 0.98 | 0.93 | 0.96 |
|  | Fault 2 | 1.00 | 1.00 | 1.00 |
| Model 2 | Normal | 1.00 | 1.00 | 1.00 |
|  | Fault 1 | 0.88 | 0.73 | 0.80 |
|  | Fault 2 | 1.00 | 0.99 | 0.99 |

The SHAP method is implemented using Python 3.7 to quantify the feature attributions to the target prediction: fault classification. The permutation explainer is utilized for both models, taking approximately 3 hours to calculate the feature attributions for all 59,077 samples of 20 features each benchmarked on a single machine with an Intel Core i7-10750H CPU @ 2.60 GHz and 32 GB of RAM. After the SHAP values have been computed, the UMAP and HDBSCAN combination is utilized to produce clustering results, as mentioned previously. For both models, the same UMAP settings are used as in the unsupervised case, and similar HDBSCAN settings with a minimum cluster size of 20 are used so to achieve a comparable number of total clusters. **Fig. 3** illustrates the semi-supervised SHAP clustering assignment for Model 1.

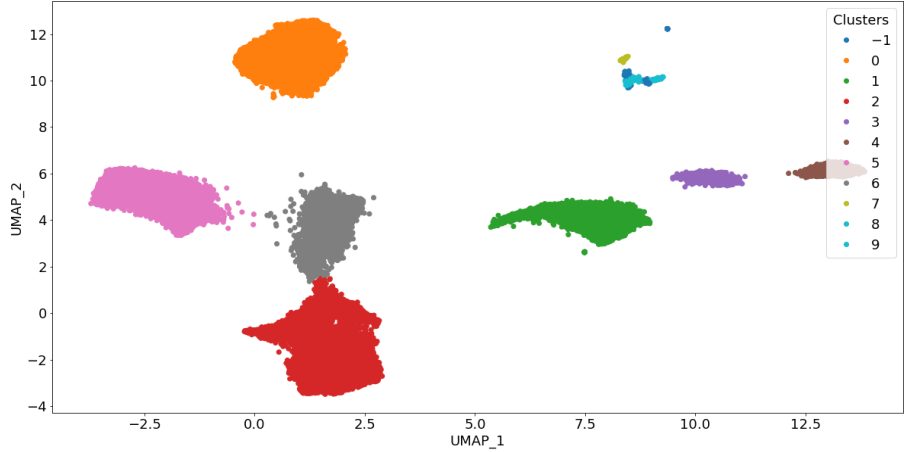

**Fig. 3** UMAP + HDBSCAN clusterings based on SHAP values for Model 1, demonstrating significantly fewer unclustered samples (~0.1%) as clusters now relate to the target prediction

The difference between the clustering results illustrated in **Fig. 2** and **Fig. 3** is stark. The additional SHAP-based transformation introduces significant structure to the clustering space, in which it becomes more evident which clusters are shaped by the various underlying model predictions. For example, the unclustered samples in addition to clusters 7, 8, and 9 from **Fig. 3** are all distinctly anomalous upon inspection; cluster 7 contains most of the Fault 2 anomalies and Fault 1 anomalies are mostly split between clusters 8 and 9. These strong associations between the clusters and faulty classes were not present in the unsupervised case illustrated in **Fig. 2**. Notably, just 0.1% of the dataset remains unclustered by HDBSCAN in this scenario, improving the clustering percentage—thereby reducing the number of designated noisy samples—by an order of magnitude compared to the purely unsupervised result. This significant improvement in clustering is made possible by having experts partially label just 1.6% of the dataset and train a supervised learning model based on those labels. The equivalent clustering assignment for Model 2 is presented in **Fig. 4.**

Once again, SHAP provides structure to the dataset that streamlines the clustering process, even when the underlying model misclassifies a significant portion of the known Fault 1 classifications. For Model 2, 0.5% of the dataset remained unclustered by HDBSCAN, still representing a significant improvement over the purely unsupervised case. Similar to the clustering obtained from Model 1, Model 2 separates most of the identified Fault 1 samples into two clusters (3 and 10). Interestingly, the NMI between the two clustering assignments is 0.86, indicating strong alignment in the information content in the clusters.

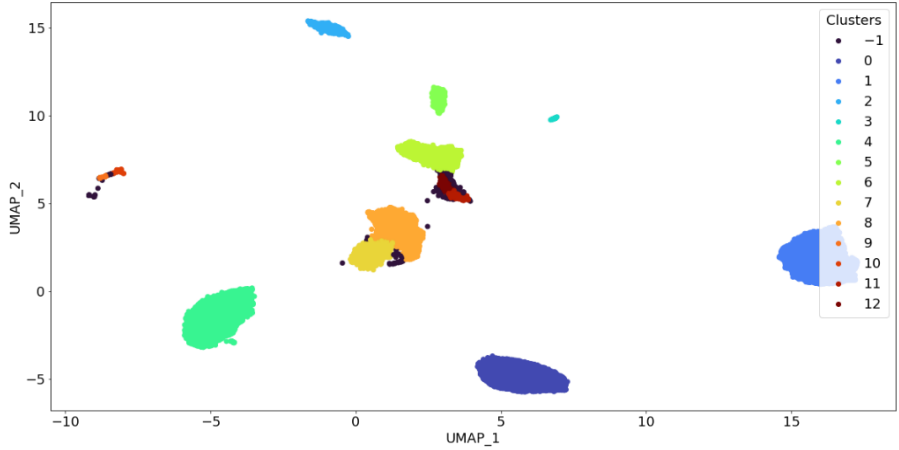

**Fig. 4** UMAP + HDBSCAN clusterings based on SHAP values for Model 2, demonstrating similar clustering quality despite differences in the underlying model performance

These findings have implications on the problem of handling imbalanced classes that are distinct from existing work. Some of the most prevalent class imbalance methods in the past have included manipulating datasets to weigh rare classes more, either by oversampling or generating more instances of rare events, or undersampling common classes so that they influence the overall prediction less (Longadge et al. 2013). The approach to utilize Shapley values to explain and visualize imbalanced class predictions is an alternative without any of these modifications. While Model 1 and Model 2 in this case study are marked by differences in classifying Fault 1, Model 2 can still provide comparable clustering assignments that are similar to Model 1 in information content, and still is a significant improvement over purely unsupervised clustering. Throughout this case study, we have demonstrated how Shapley-based analysis can be performed to explain and visualize how models make fault predictions. The next case study shifts to address the problem of predicting future faults, with Shapley-based clustering methodology utilized again to derive human-explainable and information-dense clusters.

## 4 Case Study: Turbofan Engine Prognostics

The second case study presented in this paper directs attention to a recent PHM benchmark dataset created by the NASA Prognostics Center of Excellence (PCoE) (Chao et al. 2021a) that was used for the 2021 PHM Data Challenge hosted by the PHM Society (Chao et al. 2021b). This dataset consists of time series sensor measurements from simulated turbofan engine flight cycles generated by the Commercial Modular Aero-Propulsion System Simulation (C-MAPSS) dynamical model used previously for benchmarking PHM approaches. The 2021 Challenge dataset offers realistic run-to-failure degradation trajectories for a fleet of turbofan gas engines with labeled failure mode information with component-level granularity. While the challenge itself focused on predicting the remaining useful life (RUL) of the engine units, the labeled failure modes offer an opportunity for prognostics approaches incorporating XAI techniques.

The dataset consists of 8 subsets that contain 7 different failure modes, in which each failure mode is characterized by the presence of potentially overlapping faults encountered for 5 mechanical components: fan, low pressure compressor (LPC), high pressure compressor (HPC), low pressure turbine (LPT), and high pressure turbine (HPT). Existing work on this dataset has explored deep learning techniques for RUL estimation, with challenge winners implementing variations of convolutional neural network (CNN) architectures to achieve accurate predictions (Solís-Martín et al. 2021; DeVol et al. 2021; Lövberg 2021). These approaches focused solely on RUL prediction enabled by deep feature representations, but did not forecast failing components or target interpretability for their prognostic approaches.

In our work, we focus on constructing Shapley-explainable clusters to obtain subgroups describable in terms of features directly based on the original variables of the dataset. **Table 2** presents a summary of these variables, which include dynamic operating scenario descriptions and time series sensor measurements (18 time series in total), and auxiliary variables that are held constant per cycle. We refer to the challenge formulation and documentation for more information about this benchmark dataset (Chao et al. 2021a, b).

We propose examining this benchmark dataset from multiple perspectives, building from our previous work, which introduced the reformulation of this benchmark problem for forecasting failures based on the labeled failure mode information (Cohen et al. 2023). To enhance the interpretability and prognostic utility of the data-driven model, we aim to: 1) predict the current health status, with validation possible by using the binary health state label provided by NASA; and 2) predict the failing component(s) responsible for the failure in addition to RUL prediction; and 3) explain the behavior of the predictive model by assessing feature attributions, which was not considered in any prior work. Expanding the number of outputs to a total of 7 for our model allows for the simultaneous detection of incipient faults, monitoring of equipment health, and prediction of the RUL until catastrophic failure. The prognostic insights from such an approach could allow for improved decision-making resulting in swift resource allocation and appropriate maintenance staffing, reducing costs associated with expensive reactive maintenance policies (Selcuk 2016).

**Table 2** Complete variable descriptions from the 2021 PHM Data Challenge (Chao et al. 2021b)

| Variable | Symbol | Description | Units |
|---|---|---|---|
| $A_1$ | unit | Unit number | - |
| $A_2$ | cycle | Flight cycle number | - |
| $A_3$ | $F_c$ | Flight class | - |
| $A_4$ | $h_s$ | Health state | - |
| $W_1$ | alt | Altitude | ft |
| $W_2$ | Mach | Flight Mach number | - |
| $W_3$ | TRA | Throttle-resolver angle | % |
| $W_4$ | T2 | Total temp. at fan inlet | °R |
| $Xs_1$ | Wf | Fuel flow | pps |
| $Xs_2$ | Nf | Physical fan speed | rpm |
| $Xs_3$ | Nc | Physical core speed | rpm |
| $Xs_4$ | T24 | Total temp. at LPC outlet | °R |
| $Xs_5$ | T30 | Total temp. at HPC outlet | °R |
| $Xs_6$ | T48 | Total temp. at HPT outlet | °R |
| $Xs_7$ | T50 | Total temp. at LPT outlet | °R |
| $Xs_8$ | P15 | Total pressure in bypass-duct | psia |
| $Xs_9$ | P2 | Total pressure at fan inlet | psia |
| $Xs_{10}$ | P21 | Total pressure at fan outlet | psia |
| $Xs_{11}$ | P24 | Total pressure at LPC outlet | psia |
| $Xs_{12}$ | Ps30 | Static pressure at HPC outlet | psia |
| $Xs_{13}$ | P40 | Total pressure at burner outlet | psia |
| $Xs_{14}$ | P50 | Total pressure at LPT outlet | psia |

In order to proceed with the Shapley analysis, a supervised machine learning model is required to generate predictions. To prepare this model, 7 statistical time domain features (mean, standard deviation, minimum, first quartile, median, third quartile, and maximum) are extracted from each of the 18 time series signals. The feature set is further augmented with the cycle number and flight class auxiliary variables, as well as the total time duration of the flight. For this study, an 80%-20% randomized training-testing split is utilized for hold-out testing, with 1365 cycles comprising the testing set. With 129 total features extracted per cycle, the inputs are min-max normalized to conclude the pre-processing procedure.

An ANN utilized for Shapley analysis (named xANN) is trained, learning the 7 previously specified prognostic outputs using the Flux backend (Innes 2018) in a Julia 1.7.3 computing environment. Compared to the deep learning approaches attempted previously, the xANN model is exceedingly simple, with 64 and 32 units in the two hidden layers. The training loss function is a weighted combination of binary crossentropy loss for classification outputs and NASA's asymmetrical scoring function for RUL estimation (Chao et al. 2021b), and the ReLU function is employed for intermediate activations. **Table 3** shows the classification predictions for the developed xANN model, with the most success achieved for predicting fan, HPT, and HPC failures. Notably, the simplified xANN model achieves an RMSE of 7.96 for RUL predictions with lower complexity (on the order of $10^4$ trainable parameters versus $10^6$ from prior work).

**Table 3** Health state predictions and equipment forecasts for xANN prognostics model

| Prediction | Precision | Recall | F1-Score |
|---|---|---|---|
| Unhealthy | 0.99 | 0.98 | 0.99 |
| Healthy | 0.96 | 0.98 | 0.97 |
| No Fan Failure | 0.96 | 0.96 | 0.96 |
| Fan Failure | 0.92 | 0.93 | 0.93 |
| No LPC Failure | 0.88 | 0.94 | 0.91 |
| LPC Failure | 0.85 | 0.75 | 0.80 |
| No HPC Failure | 0.94 | 0.97 | 0.96 |
| HPC Failure | 0.96 | 0.93 | 0.95 |
| No HPT Failure | 0.91 | 0.90 | 0.91 |
| HPT Failure | 0.92 | 0.92 | 0.92 |
| No LPT Failure | 0.89 | 0.77 | 0.82 |
| LPT Failure | 0.80 | 0.90 | 0.85 |

### 4.1 Shapley-based Clustering Analysis for PHM Case Study

For each of the 7 output variables, 176,085 Shapley values are approximated (1365 test samples with 129 features each) using ShapML.jl, a Julia language implementation of Štrumbelj and Kononenko's stochastic sampling algorithm (Štrumbelj and Kononenko 2014) that has been benchmarked to be more computationally efficient compared to SHAP (Redell 2020). Under the same computing environment described earlier and with a sample size setting of 60 Monte Carlo samples, it takes approximately 375 seconds to obtain the estimated Shapley values per each of the 7 output variables.

The impact of Shapley values for clustering this application is clearly illustrated in **Fig. 5**. Once again, UMAP is used for visualization purposes (McInnes 2018), with the same parameter setting as in the previous semiconductor manufacturing case study. This figure depicts the UMAP space embedded in 2 dimensions before and after the Shapley-based data transformation with respect to a single target prediction: the current health state. In **Fig. 5 a)**, healthy and unhealthy samples from the test set are not separable on the UMAP component space, dispersed with no clear pattern; on the other hand, **Fig. 5 b)** showcases the almost perfect separability of the healthy and unhealthy predicted samples present in the UMAP space following the Shapley transformation. For HDBSCAN clustering, the parameter settings are kept constant for all target predictions, including a minimum cluster size of 40 and a noise setting of 10 minimum samples.

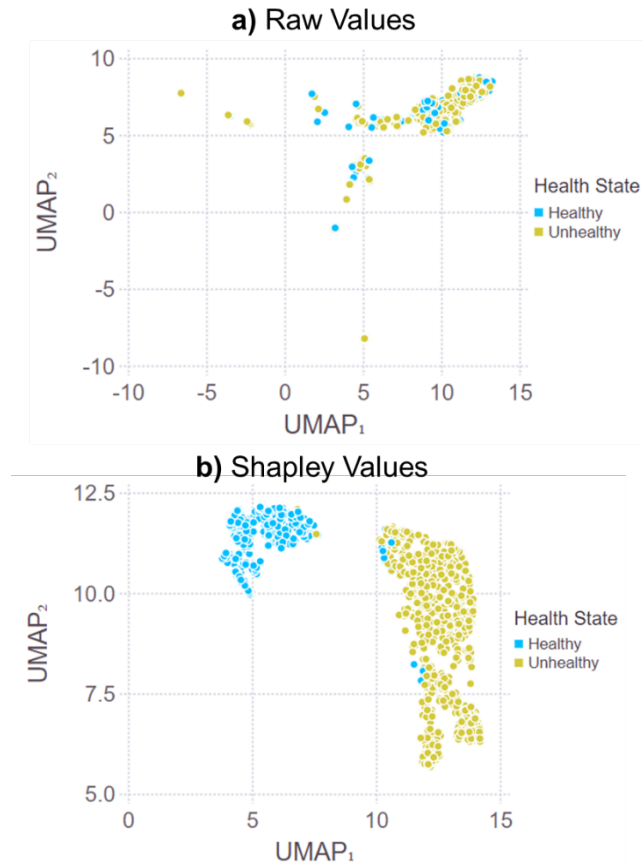

**Fig. 5** Depiction of UMAP component space colored by current health state predictions learned from **a)** raw (min-max normalized) values versus **b)** Shapley values, with the visualization in **b)** showing separability

      The Shapley-based clusters show visual separability, but most importantly can also be described using the original features. Using the SkopeRules implementation on Python 3.7, high-precision rules describe the derived clusters pictured in **Fig. 5** based on just one term: the current cycle. This is intuitive as an initial example because the engine units are healthy in initial operation (i.e., when the cycle number is low). **Fig. 6** illustrates the same clusters colored by the cycle number, where it is visually clear that the mostly healthy cluster can be described by a low cycle number.

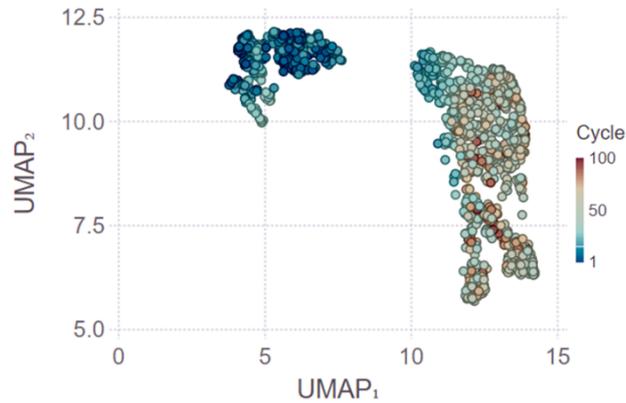

**Fig. 6** Health state clusters colored by cycle, demonstrating the ease of describing the derived Shapley-based clusters in terms of the original feature scale

The health state prediction is an intuitive case in which one variable, the current cycle, is clearly dominant in explaining the prediction. However, the other predictions of failing components are significantly more challenging due to the overlap present between failures. To clarify the subgroups of failing components, the derived clusters will consist of the failure predictions only; in other words, explainable subgroups of predicted component failures will be identified.

The Shapley-based clustering results for forecasted fan failures will be illustrated as an example of successful explainable component-level prognostic clustering. First, the top 10 features are ranked by mean absolute Shapley value, which becomes the basis for the subsequent clustering and derived descriptions. **Fig. 7** depicts the global feature importance ranking for predicting fan failures.

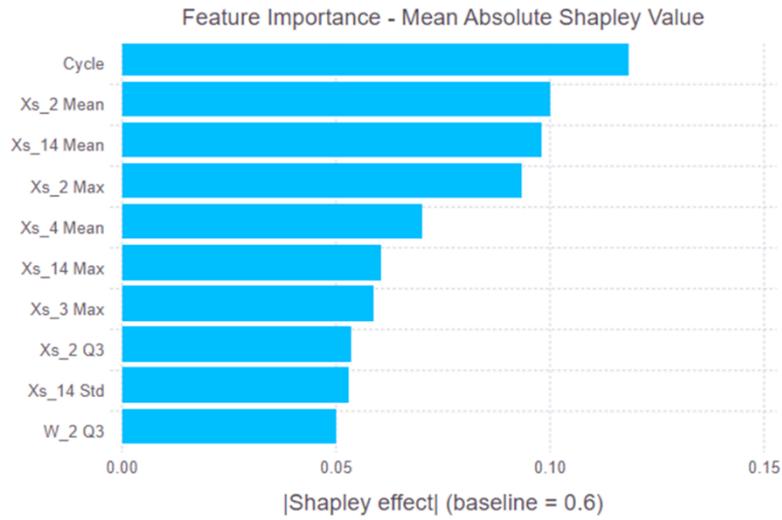

**Fig. 7** Global feature importance ranking for predicting eventual fan failures, with cycle and average physical fan speed representing the top 2 most influential features for the xANN model (see **Table 2** for complete variable descriptions)

Using the same UMAP and HDBSCAN settings as previously described, the clusters for eventual fan failures are depicted in **Fig. 8**. HDBSCAN detected two clusters (in addition to some noisy samples, removed from the **Fig. 8** illustration), with cluster 0 comprising most of the predictions.

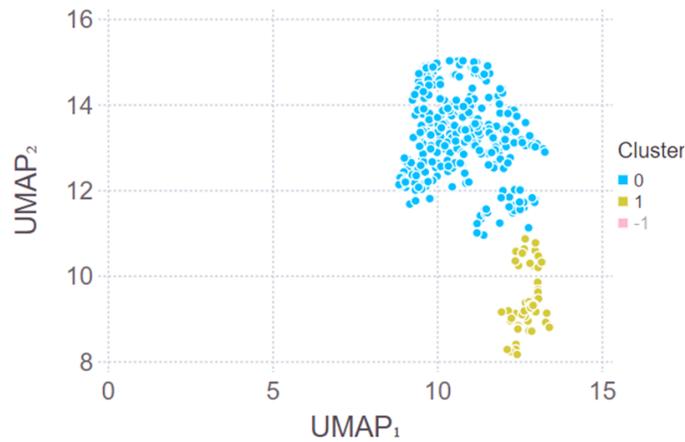

**Fig. 8** Shapley-based eventual fan failure clusters obtained by HDBSCAN, with identified noisy samples in the test set omitted from the illustration

The highest performing rule identified by SkopeRules that describes the major cluster involves the physical fan speed: both the mean and the third quartile statistics have quantifiable thresholds that describe the dominant fan failure cluster with a precision of 0.97. This procedure is followed across all forecasted component failures, with Shapley values recomputed for each target prediction. **Table 4** lists the identified rules for each of the clusters derived using the SkopeRules method across all target predictions, including health state and RUL.

**Table 4** Identified rules for Shapley-based clusters in terms of original feature values (see **Table 2** for complete variable descriptions)

| Prediction | Cluster | Identified Rule | Precision | Recall | F1-Score |
|---|---|---|---|---|---|
| Health State | 0 | Cycle ≤ 16.5 | 1.00 | 0.73 | 0.84 |
|  | 1 | Cycle > 32.5 | 1.00 | 0.83 | 0.91 |
| Fan Failure | 0 | $Xs_2$ Mean ≤ 1343.9 AND $Xs_2$ Q3 ≤ 1424.6 | 0.97 | 0.93 | 0.95 |
|  | 1 | Cycle ≤ 59.0 AND $Xs_{14}$ Mean > 2.8 | 0.72 | 0.77 | 0.74 |
| LPC Failure | 0 | $W_3$ Mean ≤ 54.2 AND $Xs_{12}$ Mean > 1863.6 | 0.90 | 0.90 | 0.90 |
|  | 1 | Cycle ≤ 65.5 AND $W_3$ Mean > 54.2 | 0.95 | 0.97 | 0.96 |
| HPC Failure | 0 | $Xs_2$ Mean ≤ 1344.6 AND $Xs_2$ Q3 > 1400.3 | 0.66 | 0.92 | 0.77 |
|  | 1 | $W_3$ Mean ≤ 59.9 AND $Xs_2$ Mean > 1339.3 | 0.77 | 0.93 | 0.84 |
|  | 2 | $Xs_{14}$ Mean ≤ 2.6 AND $Xs_4$ Mean ≤ 1146.3 | 0.96 | 0.88 | 0.92 |
| HPT Failure | 0 | $Xs_{13}$ Max > 8788.0 AND $Xs_2$ Mean > 1323.9 | 0.96 | 0.88 | 0.92 |
|  | 1 | $Xs_{13}$ Max ≤ 8788.5 AND $Xs_{13}$ Mean ≤ 8318.2 | 0.89 | 0.96 | 0.92 |
| LPT Failure | 0 | $Xs_{11}$ Q1 ≤ 10.1 AND $Xs_4$ Mean ≤ 1149.6 | 0.96 | 0.97 | 0.97 |
|  | 1 | $Xs_{14}$ Mean > 2.6 AND $Xs_4$ Mean ≤ 1149.0 | 0.53 | 0.86 | 0.65 |
|  | 2 | $Xs_{13}$ Max > 8741.3 AND $Xs_4$ Mean > 1149.1 | 0.93 | 0.91 | 0.92 |
| RUL | 0 | $W_1$ Mean ≤ 12153.6 AND $W_1$ Q3 ≤ 16356.3 | 1.00 | 0.99 | 0.99 |
|  | 1 | $W_1$ Mean > 12258.3 AND $W_1$ Q3 > 16277.0 | 1.00 | 0.98 | 0.99 |

Of the 16 total derived cluster descriptions in **Table 4**, 12 of them characterize the respective clusters with a precision exceeding 0.85 and highlight key contributions for the forecasted failures. This is particularly notable when considering that these rules are constrained to comprise a maximum of just 2 terms, using variables limited to the top 10 globally important features as quantified by mean absolute Shapley value. Some of these identified variables align with prior expectations; for example, the physical fan speed variable characterizing the major fan failure cluster. However, other variables such as the total pressure at the burner outlet explaining both forecasted HPT failure clusters are surprising findings that would have been difficult to pinpoint without XAI techniques. Another example is that the RUL clusters can be described with near-perfect separability based on the altitude, perhaps suggesting that a prognostics model calibrated or normalized to handle dynamic operating conditions could clarify the degradation trend (Lövberg 2021). These discoveries can lead to the potential identification of root failure causes, particularly if forecasted failures are closely investigated by domain experts.

## 5  Discussion

The incorporation of multiple case studies varying by level of supervision, data type, and application area shows the broad potential of XAI approaches for fault diagnosis and prognosis. The contributions in this paper revolve around the usage of Shapley-based supervised clustering, which was extended to include semi-supervised clustering with significant class imbalance. When compared to purely unsupervised clustering with the same dimensionality reduction and clustering techniques, the Shapley-based semi-supervised clustering alternative consistently resulted in improved clustering, even with imperfect underlying supervised machine learning models used. This also demonstrates the utility of having partially labeled data as opposed to a fully unlabeled dataset; even when labels are not complete or fully accurate, the targeting of specific samples by human experts enables augmented intelligence techniques to approach the problem of diagnosing exceedingly rare anomalous events.

Moreover, the development of Shapley-based XAI as benchmarked on the N-CMAPSS dataset uniquely demonstrates the potential for an explainable approach that simultaneously detects the current health state, forecasts which component(s) will fail, and then estimates the number of cycles until failure. In essence, this integrates the important disciplines of anomaly detection and fault diagnosis—conventionally requiring multiple models—in one prognostic model that makes accurate and explainable predictions, even for presently healthy units. These findings have broad economic implications beyond engine prognostics, as a similar approach could potentially be applied for other PHM applications.

There remain key limitations that should be investigated for future work. Namely, existing Shapley-based computational approaches are still prohibitively expensive for online deployment, with required resources scaling past the point of feasibility for real-time decision-making. However, the capability of understanding model predictions is key for augmented intelligence approaches that keep human experts in the loop. Additional explainability provided in terms of original feature values and describable clustering could lead to a paradigm shift for how operators interact and interface with AI systems. Under this framework, experts can explain localized predictions of interest as well as global trends with a unified approach that is a step in the direction of demystifying black-box approaches into more trustworthy and reliable "glass box" fault diagnosis models.

## 6 Conclusion

The Shapley-based clustering approach proposed in this paper derives explainability from existing data-driven fault diagnosis and prognosis models. By extending Shapley-based clustering methodology to semi-supervised problems under the critical lens of fault diagnosis and PHM, it is possible to utilize XAI techniques for practical intelligent manufacturing problems where labeled data are difficult to obtain. The main contributions of this paper are listed as follows:

- Proposed an extension of Shapley-based clustering to include semi-supervised learning, enabling explainable analysis of heavily imbalanced, partially labeled datasets;
- Demonstrated the utility of SHAP to improve dense clustering for a semi-supervised case study in semiconductor manufacturing;
- Derived informative Shapley-based fault clusters describable with high-precision rules to extend the 2021 PHM Data Challenge, with 12 out of 16 derived rules describing the clusters with a precision exceeding 0.85.

These contributions can help streamline AI adoption in industry, as explainable methodology will improve how human experts interact with and trust machine learning models. These advancements enable tangible improvements for fault diagnosis applications utilizing human-centered augmented intelligence for advanced manufacturing.

## 7 Data Availability

The dataset for the 2021 PHM Data Challenge is publicly available from NASA's Prognostics Center of Excellence Data Set Repository, accessible for download from the following link: https://www.nasa.gov/content/prognostics-center-of-excellence-data-set-repository under the heading "17. Turbofan Engine Degradation Simulation-2". Upon publication, the authors intend to additionally include a link to a GitHub repository containing the code written for this benchmark dataset.

## 8 Declarations and Statements

### 8.1 Competing Interests

The authors have no competing interests to declare that are relevant to the content of this article. In addition, no funding was received to assist with the preparation of this manuscript.